# Towards AI-enhanced control: a numerical technique for trajectory smoothing of a parallel robot for pancreatic surgery

Iosif Birlescu[1[0000-0002-4026-2318]], Alexandru Pusca[1[0000-0002-5804-575X]], Bogdan Gherman[1[0000-0002-4427-6231]], Calin Vaida[1[0000-0003-2822-9790]], Ionut Zima[1[0009-0007-0483-7691]], Damien Chablat[1,2[0000-0001-7847-6162]], Doina Pisla[1,3[0000-0001-7014-9431]]

[1] CESTER, Technical University of Cluj-Napoca, 400114 Cluj-Napoca, Romania
[2] École Centrale Nantes, Nantes Université, CNRS, LS2N, UMR 6004, F-44000 Nantes, France
[3] Technical Sciences Academy of Romania, B-dul Dacia, 26, 030167 Bucharest, Romania
doina.pisla@mep.utcluj.ro

**Abstract.** The paper presents a numerical approach for the end-effector trajectory smoothing of a parallel robot designed for minimally invasive pancreatic surgery. The approach is tailored for real-time master-slave control architecture and uses a 3D space mouse for command input for velocity control. The trajectory smoothing is achieved by generating S-curves in the end-effector velocity fields, thus controlling the accelerations, which in turn reduces tissue trauma in the minimally invasive procedures. Real-time control is enabled by segmenting the S-curves based on the command inputs from the 3D space mouse. A special case is considered where the acceleration time is constant for all command inputs. Numeric results demonstrate stable transitions (without abrupt changes) in both the end-effector parameter space and in the active joints parameters, thereby validating the proposed approach. Further work aims to test the approach on an experimental model and integrate it into AI-based training modules.

**Keywords:** Trajectory Smoothing, Parallel Robot, Minimally Invasive Surgery.

## 1 Introduction

Pancreatic cancer is one of the most lethal forms of cancer, with a 5-year survival rate below 10%, making it the tenth leading cause of cancer-related deaths globally [1]. The survival rate remains poor due to late diagnosis, the aggressive nature of the disease, and the complexity of its treatment. Surgery remains the only curative option in pancreatic cancer [1], therefore improving surgical outcomes is essential to increase the survival rate. Advances in robotic systems have shown promising results, providing support in complex surgical steps that often lead to complications [2].

One defining characteristic of minimally invasive surgery (MIS) robots is the presence of a Remote Center of Motion (RCM) [3], which constrains the instrument motion around the insertion point in the operating field, a concept similar to the Remote Center of Compliance [4]. A recently developed parallel robot for pancreatic MIS [3] was designed to assist with various tasks such as tissue manipulation and intraoperative imaging, reducing the need for a second surgeon in the operating room. The higher-order kinematics of the novel MIS parallel robot was studied in [3], establishing efficient models for both the inverse and forward kinematics; the Inverse Kinematic Models are further used in this paper as inputs in a case study.

Sudden, high acceleration movements of the surgical instruments during robotic MIS can cause tissue damage and other complications. Trajectory smoothing can be implemented in MIS robotic systems to avoid these effects and ensure overall improved performance during robot operation (e.g., lower inertias, less vibrations). Trajectory smoothing is a well-established topic in robotics and can be achieved based on various mathematical techniques, such as S-curves [5-9], polynomial profiles [10,11], convolution [12]. While most trajectory smoothing techniques found in the scientific literature are based on symbolic calculus, only a few of them use numerical approaches suitable for real-time command and control. One numerical approach has already proven effective for smoothing pre-planned trajectories [13].

This paper proposes a numerical trajectory smoothing approach for real-time master slave control of a MIS parallel robot, based on methods described in [13]. A special case is considered in which the input commands are received via a 3D space mouse (considered a master console), and the acceleration times (for the end-effector) are kept constant. The main contributions of the paper are as follows: (1) A numerical trajectory smoothing method using S-curves velocity profiles, adapted for real-time master-slave control; (2) Application of the proposed method to the Athena surgical parallel robot.

The applicability of the proposed method ranges from industrial to medical robots, and even training software modules, where AI can help the surgeon in selecting suitable values for specific parameters (input device sampling time, max acceleration, and jerk) to increase performance and ensure accuracy and safety. In addition, it was shown that the method reduces the control effort [13], facilitating the task of a regulator (e.g., PID) to follow a reference trajectory.

The paper is structured as follows: Section 2 describes the kinematics of the parallel robot for pancreatic MIS; Section 3 describes the proposed trajectory smoothing approach; Section 4 presents numerical results and a short discussion; and Section 5 presents the conclusions.

## 2 Kinematics of the parallel surgical robot

Figure 1 illustrates the kinematic scheme of the Athena parallel surgical robot for pancreatic MIS. A detailed analysis of the mechanism's functionality is found in [3]. The Athena robot contains the following modules:

1. An active parallel module (PM) which manipulates a surgical instrument (SI). The active joins parameters of PM are $q_i,\ i = \overline{1,3}$ .
2. A passive motion spherical module (SM) which constrains the SI in an RCM motion.

3. The surgical instrument SI of length $L$, with the actuator $q_4$ to achieve a rotation motion around the longitudinal axis of SI rod.

The RCM parameters are $[\psi, \theta, \varphi, l_{ins}]^T$ following a Z-Y-X rotation sequence and a translation along the OX axis by $l_{ins}$; note that the motion of $q_4 = \varphi$ is decoupled from the motion of PM.

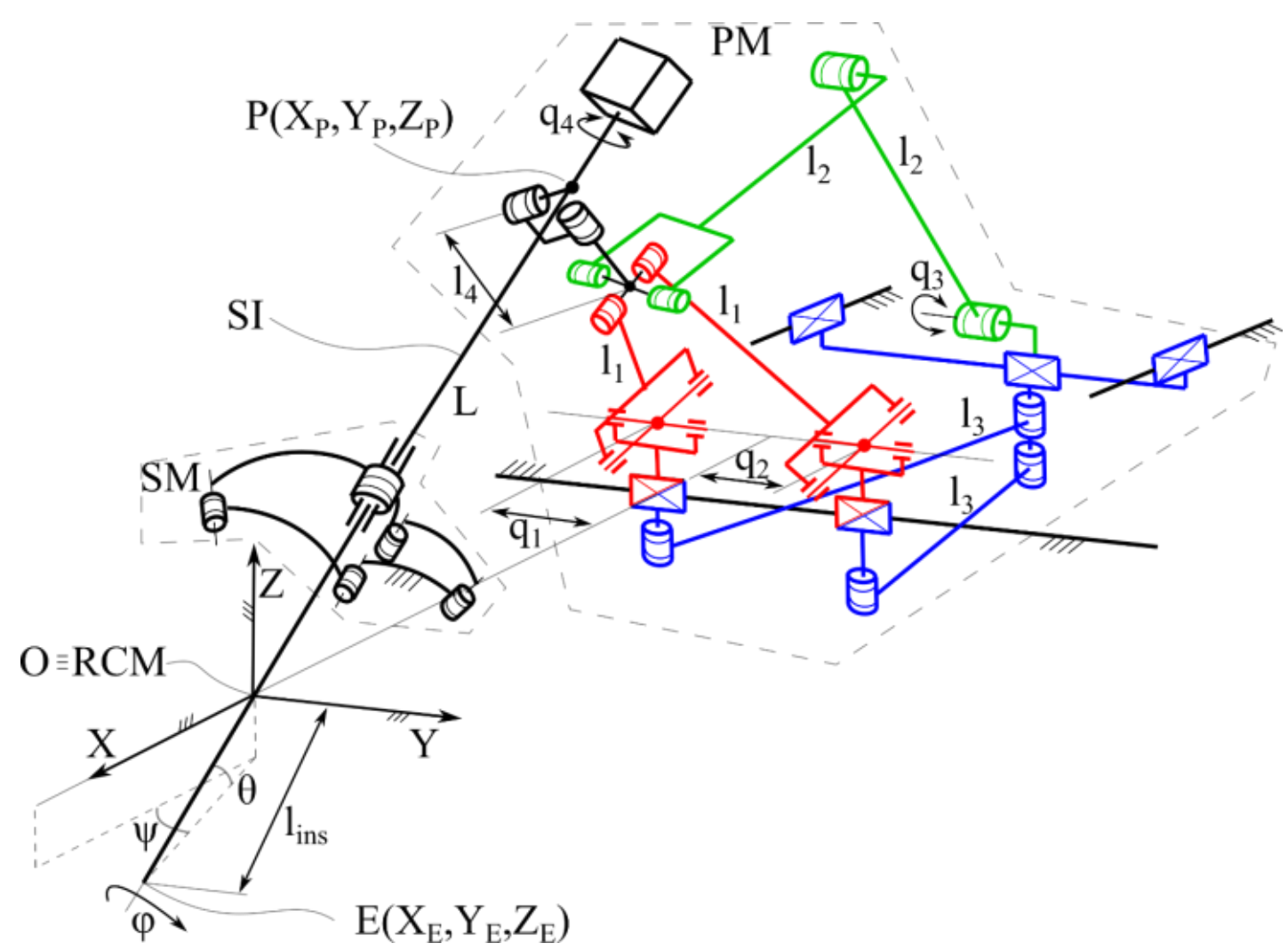

**Fig. 1.** Kinematic scheme of the parallel robot for pancreatic MIS.

For the Inverse Kinematic model, the end-effector Cartesian parameters $\mathbf{X} = [X_E, Y_E, Z_E]^T$ can be computed with respect to the RCM parameters using [3]:

$$\begin{bmatrix} X_E \\ Y_E \\ Z_E \end{bmatrix} = \begin{bmatrix} l_{ins}\cos(\psi)\cos(\theta) \\ l_{ins}\sin(\psi)\cos(\theta) \\ -l_{ins}\sin(\theta) \end{bmatrix}. \tag{1}$$

In a master-slave control approach, where surgical instrument tip parameters $\mathbf{X} = [X_E, Y_E, Z_E]^T$ are inputs, the Cartesian coordinates of $\mathbf{P} = [X_P, Y_P, Z_P]$ can be computed using:

$$\begin{bmatrix} X_P \\ Y_P \\ Z_P \end{bmatrix} = \begin{bmatrix} -X_E\sigma \\ -Y_E\sigma \\ -Z_E\sigma \end{bmatrix}, \; \sigma = \frac{(L - l_{ins})}{l_{ins}}, \; l_{ins} = \sqrt{X_E^2 + Y_E^2 + Z_E^2}. \tag{2}$$

Based on the components of **P** the following notation is used:

$$\begin{bmatrix}\rho_1\\ \rho_2\\ \rho_3\end{bmatrix} = \begin{bmatrix} Y_P \\ \sqrt{Z_P^2 + (X_P + l_0)^2} \\ \operatorname{atan2}(X_P + l_0, Z_P)\end{bmatrix}. \tag{3}$$

Furthermore, the following simultaneous equations:

$$\begin{cases} \rho_1 - \frac{1}{2}(q_1 + q_2) = 0 \\ \left(\frac{1}{2}q_1 + \frac{1}{2}q_2\right)^2 + \left(\rho_2 - l_4\right)^2 - l_2^2 = 0 \\ \left(l_3' - l_2\sin(q_3) + l_1'\sin(\rho_3)\right)^2 + \left(l_2\cos(q_0) - l_1'\cos(\rho_3)\right)^2 - l_2^2 = 0 \end{cases}, \tag{4}$$

$$l_1' = \sqrt{l_1^2 - \left(\frac{1}{2}q_1 + \frac{1}{2}q_2\right)^2},\ l_3' = \sqrt{l_3^2 - \left(\frac{1}{2}q_1 + \frac{1}{2}q_2\right)^2},$$

can be solved for $\{q_1, q_2, q_3\}$, and substituting $\{\rho_1, \rho_2, \rho_3\}$ (Eq. 3) into the solution achieves the Inverse Kinematic Model for displacement, in closed form [3] (the working mode is assumed as illustrated in Figure 1). The solution is not presented explicitly (due to its length); however, in general form, by considering real functions $f_i : \mathbb{R}^3 \to \mathbb{R},\ i = \overline{1,3}$ the solution is:

$$\mathbf{Q} = \begin{bmatrix} q_1 \\ q_2 \\ q_3 \end{bmatrix} = \begin{bmatrix} f_1(X_E, Y_E, Z_E) \\ f_2(X_E, Y_E, Z_E) \\ f_3(X_E, Y_E, Z_E) \end{bmatrix}. \tag{5}$$

To compute the Inverse Kinematic Model for velocities and accelerations the following equations are used:

$$\begin{aligned} \dot{\mathbf{Q}} &= -\mathbf{A}\cdot\dot{\mathbf{X}}, \\ \ddot{\mathbf{Q}} &= -\left(\dot{\mathbf{A}}\cdot\dot{\mathbf{X}} + \mathbf{A}\cdot\ddot{\mathbf{X}}\right), \end{aligned} \tag{6}$$

where **A** is the Jacobian computed from $f_i,\ i = \overline{1,3}$ by differentiating with respect to the components of **X**.

## 3 Trajectory smoothing approach

For the Athena parallel robot master-slave control two approaches are foreseen, namely using the Omega 7 Haptic device, and the 3D space mouse 3DConnexion. This study focuses specifically on command inputs from a 3D space mouse (a lower-cost alternative to haptics, which is not explored in the literature). Smoothing in this study refers to controlling the acceleration and deceleration of the end-effector on an arbitrary path to ensure low inertia.

An input on a space mouse, i.e., a change of its position from the zero (neutral) position to an arbitrary position will be recorded in a vector $\mathbf{V} = [X_V, Y_V, Z_V]$ with $X_V, Y_V, Z_V \in [0,1]$, for every time instance $T \in \mathbf{T} = [T_0, T_1, \ldots, T_n]$, subject to the sampling time $T_S = T_{i+1} - T_i$ (for the space mouse input read) or a sampling frequency of $T_S^{-1}$. The vector $\mathbf{V}$ contains only Cartesian components (the space mouse orientations are disabled) since the command of the end-effector is done for the Cartesian coordinates $\mathbf{X} = [X_E, Y_E, Z_E]$; $\mathbf{V}$ can be viewed as a velocity input for the Tool Center Point (TCP) at every time instance $T$.

***Definition 1***. *The sets:*

$$\Delta_X = \{0 = x_0 < x_1 < x_2 \ldots < x_n \le 1\},$$
$$\Delta_Y = \{0 = y_0 < y_1 < y_2 \ldots < y_n \le 1\},$$
$$\Delta_Z = \{0 = z_0 < z_1 < z_2 \ldots < z_n \le 1\},$$

*are the partitions of the domains associated with* $X_V, Y_V, Z_V$*, respectively. Then, the thresholds:*

$$v_X = \left\{ v_{Xj} \mid x_{j-1} < X_V < x_{j+1},\ j = \overline{1, n-1},\ v_{X1} < v_{X2} < \ldots < v_{Xn} \right\},$$
$$v_Y = \left\{ v_{Yj} \mid y_{j-1} < Y_V < y_{j+1},\ j = \overline{1, n-1},\ v_{Y1} < v_{Y2} < \ldots < v_{Yn} \right\},$$
$$v_Z = \left\{ v_{Zj} \mid z_{j-1} < Z_V < z_{j+1},\ j = \overline{1, n-1},\ v_{Z1} < v_{Z2} < \ldots < v_{Zn} \right\},$$

*are the discrete mappings of the space mouse input to the velocity command for the TCP. Associating* $v_X, v_Y, v_Z$ *with the time vector* $\mathbf{T} = [T_0, T_1, \ldots, T_n]$ *yields:*

$$v_X(T) = [v_{Xa}(T_i)], v_Y(T) = [v_{Yb}(T_i)],\ v_Z(T) = [v_{Zc}(T_i)],$$
$$a, b, c \in [1, n] \subset \text{¥},\ i = \overline{1, n}.$$

A smoothed trajectory was defined in [13] as $\mathbf{\Gamma}^{[k]}(t)$ where $k$ is the degree of smoothing. For the MIS parallel robot, the proposed smoothing will be performed in the velocity field, which means that a degree $k = 1$ for $\mathbf{\Gamma}^{[k]}(t)$ is sufficient to control the accelerations [13].

**Note 1.** *A special case is considered in this study, where the acceleration time* $\Delta t > T_s$*, and is constant for every velocity profile; this ensures a balance between responsiveness (changes in velocities happen in same time intervals) and smoothness.*

In addition, in [13], the smoothed trajectory was pre-planned. However, in the context of real-time master-slave control a segmented trajectory must be considered, with each segment having a time length of $T_S$.

***Definition 2****.* $\mathbf{\Gamma}^{[1]}(t) = [\mathbf{X}^{[1]}(t),\ \mathbf{Y}^{[1]}(t),\ \mathbf{Z}^{[1]}(t)]^{\mathrm{T}} = [\mathbf{\Gamma}_1^{[1]}(t),\ \mathbf{\Gamma}_2^{[1]}(t), \ldots, \mathbf{\Gamma}_n^{[1]}(t)]^{\mathrm{T}}$ *is a discrete smoothed trajectory of degree* 1*, where* $\mathbf{\Gamma}_i^{[1]}(t) = [\mathbf{X}_i^{[1]}(\mathbf{t}_i),\ \mathbf{Y}_i^{[1]}(\mathbf{t}_i),\ \mathbf{Z}_i^{[1]}(\mathbf{t}_i)]^{\mathrm{T}}$ *is a segment of* $\mathbf{\Gamma}^{[1]}(t)$ *; the time vectors* $\mathbf{t}_1 = [0, \ldots, \Delta t],\ \mathbf{t}_i = \mathbf{t}_{i-1} + \Delta t,\ i = \overline{1, n}$ *have a sampling time* $t_S$ *with the constraint that* $T_S = u \cdot t_S,\ u \in ¥^*$ *. The discrete functions of the end-effector parameters* $\mathbf{X}^{[1]}(t), \mathbf{Y}^{[1]}(t),\ \mathbf{Z}^{[1]}(t)$ *are of class* $C^1$ *.*

Two cases that are considered based on the space mouse inputs:

1. Computing smooth trajectories for a single input change in $\mathbf{v}$ during $\Delta t$ . For this case the inputs are $\{v_i, v_f, a_{\max}, \Delta t\}$ (initial and final velocity, maximum acceleration and the time of acceleration); the transient time parameters are computed using [13]:

$$\Delta v = v_f - v_i, t_{a_ct} = \frac{2 \cdot \Delta v - \Delta t \cdot a_{\max}}{a_{\max}}, t_{ramp_u} = t_{ramp_d} = \frac{\Delta t - t_{a_ct}}{2}, \tag{7}$$

   where $t_{a_ct}$ is the time at constant acceleration, $t_{r_u}$ is the acceleration time and $t_{r_d}$ the deceleration time.

2. Computing smooth trajectories for multiple input changes in $\mathbf{v}$ during $\Delta t$ . For this case the inputs are $\{v_i, v_f, a_0, a_{\max}, \Delta t\}$ ( $a_0$ being the acceleration at $T$ of input change); the transient time parameters are computed using:

$$\begin{gathered} t_{r_u} = \frac{2 \cdot (a_0 - a_{\max})(\Delta v - \Delta t \cdot a_{\max})}{a_0^2 - 2 \cdot a_0 \cdot a_{\max} + 2 \cdot a_{\max}^2},\ t_{a_ct} = \frac{\Delta t\left(a_0^2 - 2 \cdot a_{\max}^2\right) - 2\Delta v\left(a_0 - 2 \cdot a_{\max}\right)}{a_0^2 - 2 \cdot a_0 \cdot a_{\max} + 2 \cdot a_{\max}^2}, \\ t^* = \frac{2 \cdot a_0\left(\Delta t \cdot a_{\max} - \Delta v\right)}{a_0^2 - 2 \cdot a_0 \cdot a_{\max} + 2 \cdot a_{\max}^2},\ t_{r_d} = t_{ramp_u} + t^*, \end{gathered} \tag{8}$$

In both cases, segments of the velocity profile are computed with u sample points since $T_S = u \cdot t_S,\ u \in ¥^*$ . Each segment is computed based on the trapezoid acceleration profile (ramps, constant acceleration or a combination between the two), with known starting and ending time, using numerical integration (cumsumtrapz [13]). The time scaling approach [13] is not required in this study since the changes in the space mouse inputs are always considered liner functions between the threshold $v_X, v_Y, v_Z$ . However, if the input velocity is considered an arbitrary function, then the time scaling normalization must be applied.

**Note 2**. *An input delay is introduced* $d_I \geq T_S$ *since any change of* $\mathbf{v} = [X_V, Y_V, Z_V]$ *can be recorded synchronously at instances of* $T \in \mathbf{T} = [T_0, T_1, \ldots, T_n]$ *. However,* $d_I$ *can be estimated (based on* $T_S$ *, PLC frequency, smoothing algorithm benchmark) and kept under values that are not noticeable by surgeons.*

Figure 2 illustrates the proposed velocity smoothing approach. Figure 2.a shows a velocity profile where the inputs ($v_1$ and $v_2$) are received at time instances greater than $\Delta t$. The input trajectory (black curves) is composed of continuous (but not smooth) linear functions (split on intervals); the smoothing approach computes the S-curves segments (green-dashed curves). Figure 2.b shows the trapezoidal profiles in the acceleration field, that produced the S-curves in the velocity field (within Figure 2.a). Figure 2.c shows a velocity profile where the inputs ($v_1$ and $v_2$) are received at time instances smaller than $\Delta t$. Figure 2.d shows the acceleration profiles used to compute the smooth velocity curves in Figure 2.c.

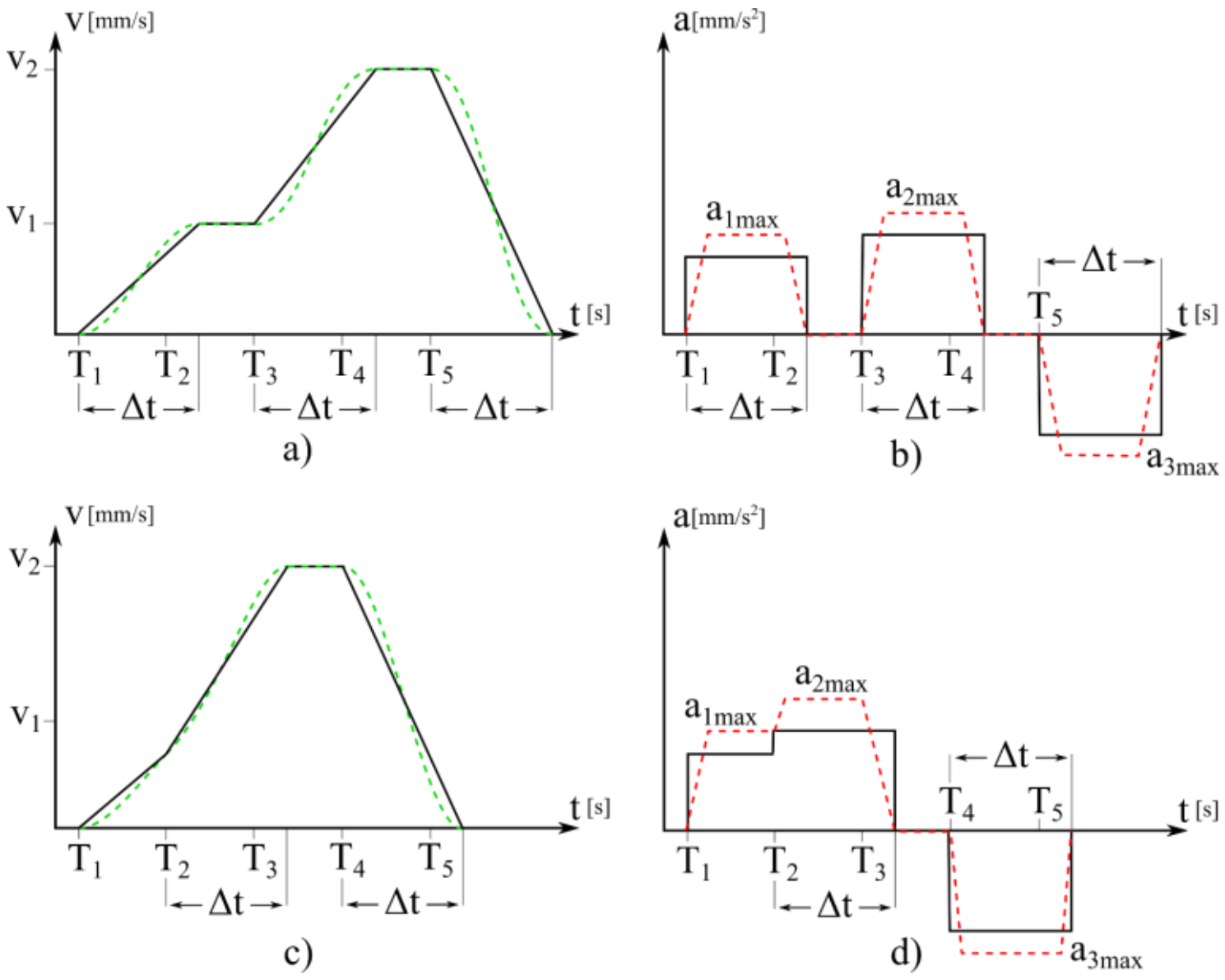

**Fig. 2.** Velocity smoothing approach (black curves – nonsmoothed, green-dashed and red-dashed curves – smoothed): a) velocity profile for a single input within Δ*t*; b) acceleration profile for a single input within Δ*t*; d) velocity profile for double input within Δ*t*; d) acceleration profile for double input within Δ*t*.

## 4 Numerical results and discussion

As a case study, the Athena parallel robot (Figure 1) is used with the following geometric parameters $\{l_0 = 300, l_1 = 200, l_2 = 150, l_3 = 170, l_4 = 50, L = 400\}\ [mm]$. For the master-slave control trajectory smoothing, the following parameters were used $\{\Delta t = 0.25,\ t_s = 0.01,\ T_S = 0.1\}\ [s]$ for acceleration times and sampling periods; consequently, the discrete trajectory sampling frequency and the space mouse input frequency are $\{t_s^{-1} = 100,\ T_S^{-1} = 10\}\ [Hz]$. Furthermore, two velocity thresholds were defined $\{v_1 = 5, v_2 = 10\}[\frac{mm}{s}]$ with the associated maximum accelerations $\{a_{1\max} = 30,\ a_{2\max} = 60\}[\frac{mm}{s^2}]$. The numerical simulation did not assume any joint displacement, velocity and acceleration limits as in [13].

An end-effector precise motion, within the surgical parallel robot workspace, was simulated based on input changes (of the space mouse) at times *T*; the initial end-effector position was $\mathbf{X}_0 = [X_E = 200, Y_E = 150, Z_E = -100][mm]$. Table 1 shows eight stages of the simulated trajectory. Figure 3 shows the time history diagrams of the end-effector Cartesian parameters $\mathbf{X} = [X_E, Y_E, Z_E]$ in displacement (green curves), velocity (blue curves), and acceleration (red curves) for the eight stages defined in Table 1; Figure 4 shows the active joint parameters $q_i,\ i = \overline{1,3}$ computed via the Inverse Kinematic models presented in Eq. (5), (6) required to follow the smoothed trajectory. The trajectory was tested to be singularity-free, meaning that $\det(\mathbf{A}) \neq 0$ (Eq. 6) and $\mathrm{sgn}(\det(\mathbf{A})) = ct$ for every discrete point on the trajectory.

**Table 1.** End-effector motion stages based on the space mouse inputs.

| *Stage.* | 1 | 2 | 3 | 4 | 5 | 6 | 7 | 8 |
|---|---|---|---|---|---|---|---|---|
| *T* [s] | 0.1 | 0.4 | 0.7 | 1 | 1.4 | 1.8 | 1.9 | 2.3 |
| $[v_X,v_Y,v_Z]$ [mm/s$^2$] | $[v_1,0,0]$ | $[v_1,0,0]$ | [0,0,0] | $[0,v_1,0]$ | [0,0,0] | $[0,0,v_1]$ | $[0,0,v_2]$ | [0,0,0] |

**Fig. 3.** Time history diagrams for the parallel MIS robot end-effector parameters (green – displacement, blue – velocity, red - acceleration).

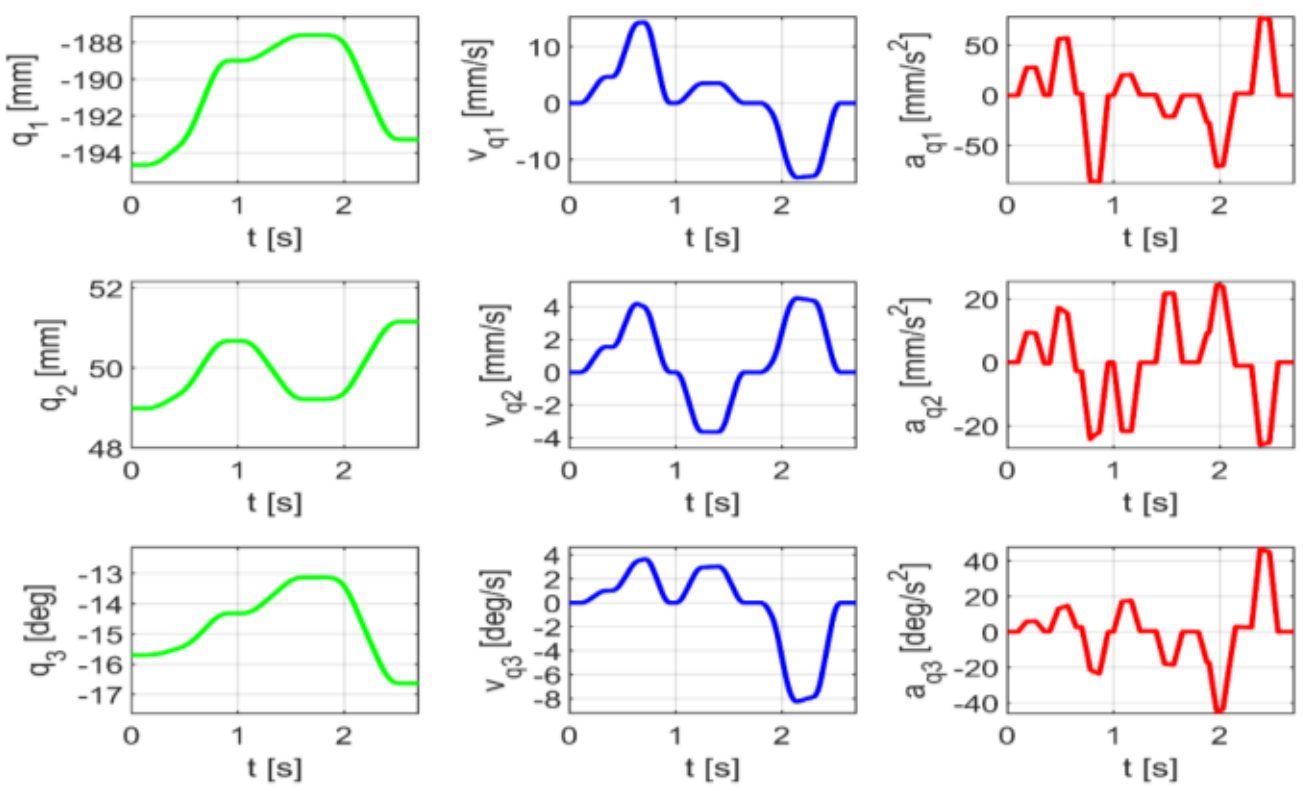

**Fig. 4.** Time history diagrams for the parallel MIS robot active joint parameters (green – displacement, blue – velocity, red - acceleration).

Regarding efficiency, as seen in Figures 3 and 4, there are no abrupt changes (spikes) in the velocity and acceleration fields for the active joint parameters. This suggests that this approach can provide good control stability and reduce the control effort, which in turn benefits the accuracy by eliminating vibrations.

Regarding operation safety during MIS, the proposed method, coupled with the time scaling approach presented in [13], can be used for automatically diminishing the velocities of the surgical instrument near the boundaries of the parallel robot workspace. This implementation aims to reduce the control effort of regulators. Similarly, the velocities of the active joints can be diminished at the joint limits or when joint limit velocities and accelerations are approached, as shown in [14].

Regarding applicability, it is reasonable to assume that various other medical robots can benefit from the proposed smoothing algorithm. In the master-slave control for needle insertion oncologic robots [15,16], the benefit comes from positioning and inserting the needle even on nonlinear trajectories. The benefit of implementing the algorithm on rehabilitation robots (for pre-planned trajectories) was discussed in [13]. In addition, the proposed method can be deployed on industrial robots or reconfigurable ones [17].

Future development of control systems and training modules will benefit from AI to personalize kinematic parameters, such as maximum acceleration, jerk limits, and sampling frequencies for the space mouse. The task of the AI in a training module is to systematically vary these parameters while monitoring the surgeon’s performance. Using this performance data, the AI system can learn optimal parameter configurations tailored to individual surgeons. This adaptive approach aims to enhance both safety and precision during robotic-assisted procedures by aligning control sensitivity with user expertise.

## 5 Conclusions

The paper proposed a numerical trajectory smoothing technique for real-time master-slave control of a parallel robot designed for minimally invasive pancreatic surgery. The method uses S-curve velocity (segmented) profiles derived from discrete inputs from a 3D space mouse.

Numerical simulations prove stable behavior in both end-effector and active joint parameters, validating the effectiveness of the proposed smoothing strategy.

Future work will focus on implementing the method in an experimental setup and validating its responsiveness and accuracy. Additionally, the system will be extended to support haptic input devices and integrated with AI-based training modules. These modules aim to personalize kinematic control parameters based on surgeon-specific data, enhancing usability and safety through adaptive control.

**Acknowledgement.**

This work was supported by the project the project *"Romanian Hub for Artificial Intelligence—HRIA"*, 428 Smart Growth, Digitization and Financial Instruments Program, MySMIS no. 334906.